\documentclass[runningheads]{llncs}
\pdfoutput=1
\usepackage{breqn}
\usepackage{hyperref}
\usepackage{color, colortbl}
\usepackage{xspace}
\usepackage{graphicx}
\usepackage[inline]{enumitem}
\usepackage{amsmath}
\usepackage{amsfonts}
\usepackage[noend,linesnumbered,lined,ruled,commentsnumbered]{algorithm2e}
\usepackage{hyphenat}
\usepackage{tabularx}
\usepackage{booktabs}
\usepackage{cleveref}
\usepackage[skip=0pt, indent=12pt]{parskip}
\newcommand{\fancyname}{\textsf{Propulate}\xspace}
\let\oldnl\nl
\newcommand{\nonl}{\renewcommand{\nl}{\let\nl\oldnl}}

\begin{document}

\title{Massively Parallel Genetic Optimization through Asynchronous Propagation of Populations}
\author{%
    Oskar~Taubert\orcidID{0000-0002-3707-499X},
    Marie~Weiel\orcidID{0000-0001-9648-4385},
    Daniel~Coquelin\orcidID{0000-0001-8552-5153},
    Anis~Farshian\orcidID{0000-0002-9888-0653},
    Charlotte~Debus\orcidID{0000-0002-7156-2022},
    Alexander~Schug\orcidID{0000-0002-0534-502X},
    Achim~Streit\orcidID{0000-0002-5065-469X}, \and
    Markus~Götz\orcidID{0000-0002-2233-1041}
}

\authorrunning{O. Taubert et al.}
\titlerunning{Propulate}
\institute{Steinbuch Centre for Computing (SCC), Karlsruhe Institute of Technology (KIT), 76344 Eggenstein-Leopoldshafen, Germany, \email{markus.goetz@kit.edu}}

\maketitle

\begin{abstract}
We present \fancyname, an evolutionary optimization algorithm and software package for global optimization and in particular hyperparameter search.
For efficient use of HPC resources, \fancyname omits the synchronization after each generation as done in conventional genetic algorithms. 
Instead, it steers the search with the complete population present at time of breeding new individuals.
We provide an MPI\hyp based implementation of our algorithm, which features variants of selection, mutation, crossover, and migration and is easy to extend with custom functionality. 
We compare \fancyname to the established optimization tool \texttt{Optuna}. 
We find that \fancyname is up to three orders of magnitude faster without sacrificing solution accuracy, demonstrating the efficiency and efficacy of our lazy synchronization approach. 
Code and documentation are available at \href{https://github.com/Helmholtz-AI-Energy/propulate}{https://github.com/Helmholtz-AI-Energy/propulate}.
\end{abstract}


\section{Introduction}
\label{sec:introduction}
Machine learning (ML) algorithms are heavily used in almost every area of human life today, from medical diagnosis and critical infrastructure to transportation and food production. 
Almost all ML algorithms have non\hyp learnable hyperparameters (HPs) that influence the training and in particular their predictive capacity. 
As evaluating a set of HPs involves at least a partial training, state\hyp free approaches to HP optimization (HPO), like grid and random search, often go beyond available compute resources~\cite{feurer2019hyperparameter}. 
To explore the high\hyp dimensional HP spaces efficiently, information from previous evaluations must be leveraged to guide the search. 
Such state\hyp dependent strategies minimize the number of evaluations to find a useful model, reducing search times and thus the energy consumption of the computation. 
Bayesian and bio\hyp inspired optimizers are the most popular of these AutoML approaches. 
Among the latter, genetic algorithms (GAs) are versatile metaheuristics inspired by natural evolution.
To solve a search\hyp for\hyp solutions problem, a population of candidate solutions (or individuals) is evolved in an iterative interplay of selection and variation~\cite{holland1992adaptation,mitchell1998introduction}. 
Although reaching the global optimum is not guaranteed, GAs often find near\hyp optimal solutions with less computational effort than classical optimizers~\cite{blum2003metaheuristics,bianchi2009survey}. 
They have become popular for various optimization problems, including HPO for ML and neural architecture search (NAS)~\cite{elsken2019neural}. 

To take full advantage of the increasingly bigger models and datasets, designing scalable algorithms for high performance computing (HPC) has become a must~\cite{young2015optimizing}. 
While Bayesian optimization is inherently serial, the structure of GAs renders them suitable for parallelization~\cite{sudholt2015pea}:
Since all candidates in each iteration are independent, they can be evaluated in parallel.
To breed the next generation, however, the previous one has to be completed.
As the computational expenses for evaluating different candidates vary, synchronizing the parallel evolutionary process affects the scalability by introducing a substantial bottleneck. 
Approaches to reducing the overall communication in parallel GAs like the island model (IM)~\cite{sudholt2015pea} do not address the underlying synchronization problem. 

To solve the issues arising from explicit synchronization, we introduce \fancyname, a massively parallel genetic optimizer with asynchronous propagation of populations and migration. 
Unlike classical GAs, \fancyname maintains a continuous population of already evaluated individuals with a softened notion of the typically strictly separated, discrete generations. 
Our contributions include:
\begin{itemize}
    \item A novel parallel genetic algorithm based on a fully asynchronous island model with independently processing workers, allowing to parallelize the optimization process and distribute the internal evaluation of the objective function.
    \item Massive parallelism by asynchronous propagation of continuous populations and migration.
    \item A prototypical implementation in Python using extremely efficient communication via the message passing interface (MPI).
    \item Optimal use of parallel hardware by minimizing idle times in HPC systems.
\end{itemize}
We use \fancyname to optimize various benchmark functions and the HPs of a deep neural network on a supercomputer. 
Comparing our results to those of the popular HPO package \texttt{Optuna}, we find that \fancyname is consistently drastically faster without sacrificing solution accuracy. 
We further show that \fancyname scales well to at least 100 processing elements (PEs) without relevant loss of efficiency, demonstrating the efficacy of our asynchronous evolutionary approach.
\section{Related Work}
\label{sec:related-work}
Recent progress in ML has triggered heavy use of these techniques with Python as the de facto standard programming language. 
Tuning HPs requires solving high\hyp dimensional optimization problems with ML algorithms as black boxes and model performance metrics as objective functions (OFs). 
Most common are Bayesian optimizers (e.g. \texttt{Optuna}~\cite{akiba2019optuna}, \texttt{Hyperopt}~\cite{bergstra2013making}, \texttt{SMAC3}~\cite{hutter2011sequential,lindauer2022smac3}, \texttt{Spearmint}~\cite{snoek2012practical}, \texttt{GPyOpt}~\cite{gpyopt2016}, and \texttt{MOE}~\cite{wang2020parallel}) and bio\hyp inspired methods such as swarm\hyp based (e.g. \texttt{FLAPS}~\cite{weiel2021dynamic}) and evolutionary (e.g. \texttt{DEAP}~\cite{fortin2012deap}, \texttt{MENNDL}~\cite{young2015optimizing}) algorithms. 
Below, we provide an overview of popular HP optimizers in Python, with a focus on state\hyp dependent parallel algorithms and implementations. 
A theoretical overview of parallel GAs can be found in surveys~\cite{cantupaz1998survey,alba1999ASO,alba2002parallelism} and books~\cite{tomassini2006spatially,luque2011parallel}. 

\texttt{Optuna} adopts various algorithms for HP sampling and pruning of unpromising trials, including tree\hyp structured Parzen estimators (TPEs), Gaussian processes, and covariance matrix adaption evolution strategy.
It enables parallel runs via a relational database server.
In the parallel case, an \texttt{Optuna} candidate obtains information about previous candidates from and stores results to disk.

\texttt{SMAC3} (\textbf{S}equential \textbf{M}odel\hyp based \textbf{A}lgorithm \textbf{C}onfiguration) combines a random\hyp forest based Bayesian approach with an aggressive racing mechanism~\cite{hutter2011sequential}. 
Its parallel variant \texttt{pSMAC} uses multiple collaborating \texttt{SMAC3} runs which share their evaluations through the file system. 

\texttt{Spearmint}, \texttt{GPyOpt}, and \texttt{MOE} are Gaussian\hyp process based Bayesian optimizers. 
\texttt{Spearmint} enables distributed HPO via Sun Grid Engine and MongoDB. 
\texttt{GPyOpt} is integrated into the \texttt{Sherpa} package~\cite{hertel2018sherpa}, which provides implementations of recent HP optimizers along with the infrastructure to run them in parallel via a grid engine and a database server. 
\texttt{MOE} (\textbf{M}etric \textbf{O}ptimization \textbf{E}ngine) uses a one\hyp step Bayes\hyp optimal algorithm to maximize the multi\hyp points expected improvement in a parallel setting~\cite{wang2020parallel}. 
Using a REST\hyp based client\hyp server model, it enables multi\hyp level parallelism by distributing each evaluation and running multiple evaluations at a time. 

\texttt{Nevergrad}~\cite{nevergrad} and \texttt{Autotune}~\cite{koch2018autotune} provide gradient\hyp free and evolutionary optimizers, including Bayesian, particle swarm, and one\hyp shot optimization. 
In \texttt{Nevergrad}, parallel evaluations use several workers via an executor from Python's \texttt{concurrent} module. 
\texttt{Autotune} enables concurrent global and local searches, cross\hyp method sharing of evaluations, method hybridization, and multi\hyp level parallelism. 
\texttt{Open Source Vizier}~\cite{song2022open} is a Python interface for Google's HPO service Vizier. 
It implements Gaussian process bandits~\cite{golovin2017google} and enables dynamic optimizer switching. 
A central database server does the algorithmic proposal work, clients perform evaluations and communicate with the server via remote procedure calls. 
\texttt{Katib}~\cite{george2020katib} is a cloud\hyp native AutoML project based on the Kubernetes container orchestration system. 
It integrates with \texttt{Optuna} and \texttt{Hyperopt}. 
\texttt{Tune}~\cite{liaw2018tune} is built on the \texttt{Ray} distributed computing platform. 
It interfaces with \texttt{Optuna}, \texttt{Hyperopt}, and \texttt{Nevergrad} and leverages multi\hyp level parallelism. 

\texttt{DEAP} (\textbf{D}istributed \textbf{E}volutionary \textbf{A}lgorithms in \textbf{P}ython)~\cite{fortin2012deap} implements general GAs, evolution strategies, multi\hyp objective optimization, and co\hyp evolution of multi\hyp populations. 
It enables parallelization via Python's \texttt{multiprocessing} or \texttt{SCOOP} module. 
\texttt{EvoTorch}~\cite{EvoTorch} is built on \texttt{PyTorch} and implements distribution- and population\hyp based algorithms. 
Using a \texttt{Ray} cluster, it can scale over multiple CPUs, GPUs, and computers. 
\texttt{MENNDL} (\textbf{M}ulti\hyp node \textbf{E}volutionary \textbf{N}eural \textbf{N}etworks for \textbf{D}eep \textbf{L}earning)~\cite{young2015optimizing} is a closed\hyp source MPI\hyp parallelized HP optimizer for automated network selection. 
A master node handles the genetic operations while evaluations are done on the remaining worker nodes. 
However, global synchronization hinders optimal resource utilization~\cite{young2015optimizing}.

\section{Propulate Algorithm and Implementation}
\label{sec:propulate}
\begin{algorithm}[t]
    \small
    \SetAlgoLined
    \DontPrintSemicolon 
    \SetKwInOut{Input}{Input}
    \Input{Search\hyp space limits, population size $P$, \textit{termination\_condition}, \textit{selection\_policy}, 
    \textit{crossover\_probability}, \textit{mutation\_probability}.
    }
    Initialize population \textit{pop} of $P$ individuals within search space. \;
    \While(\tcp*[f]{OPTIMIZE}){\textbf{not} termination\_condition}
    {
        Evaluate individuals in \textit{pop}.\tcp*[r]{EVALUATE}        
        Choose \textit{parents} from \textit{pop} following \textit{selection\_policy}. \tcp*[r]{SELECT}
        \ForEach(\tcp*[f]{VARY}){\textit{individual} \textbf{in} \textit{pop}}{
        \If(\tcp*[f]{RECOMBINE}){random $\leq$ crossover\_probability}{
        Recombine individuals randomly chosen from \textit{parents}.}
        \If(\tcp*[f]{MUTATE}){random $\leq$ mutation\_probability}{
        Mutate.}
        Update \textit{individual} in \textit{pop}.\;
    }
    }
    \KwResult{Best individual found (i.e., with lowest OF value for minimization).}
    \caption{\textbf{Basic GA.} 
    In each generation, the individuals are evaluated in terms of the optimization problem's OF. 
    Genetic operators propagate them to the next generation: 
    The selection operator chooses a portion of the current generation, where better individuals are usually preferred. 
    To breed new individuals, the genes of two or more parent individuals from the selected pool are manipulated. 
    While the crossover operator recombines the parents' genes, the mutation operator alters them randomly. 
    This is repeated until a stopping condition is met.
}
    \label{alg:basicGA}
\end{algorithm}
To alleviate the bottleneck inherent to synchronized parallel genetic algorithms, our massively parallel genetic optimizer \fancyname (\textsl{\textbf{prop}agate} and \textsl{pop\textbf{ulate}}) implements a fully asynchronous island model specifically designed for large\hyp scale HPC systems. 
Unlike conventional GAs, \fancyname maintains a continuous population of evaluated individuals with a softened notion of the typically strictly separated generations. 
This enables \textsl{asynchronous} evaluation, variation, propagation, and migration of individuals.

\begin{figure}[tb]
    \centering
    \includegraphics[width=1.\linewidth]{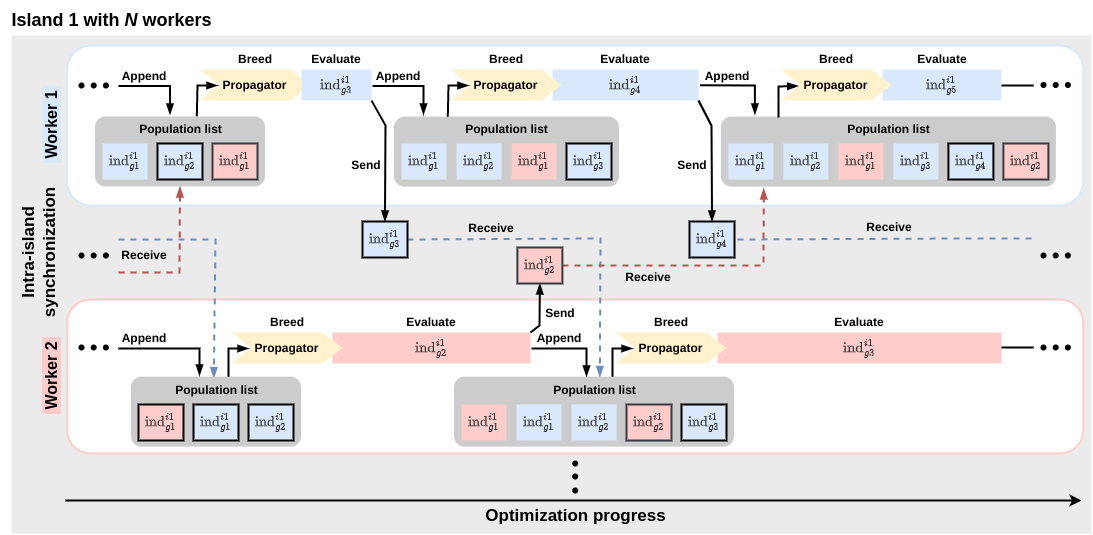}
    \setlength{\belowcaptionskip}{-10pt} 
    \caption{\textbf{Asynchronous propagation.} 
    Interaction of two workers on one island. 
    Individuals bred by worker 1 and 2 are shown in blue and red, respectively. 
    Their origins are given by a generation sub- and an island superscript. 
    Populations are depicted as round grey boxes, where most recent individuals have black outlines. 
    Varying evaluation times are represented by sharp boxes of different widths. 
    We illustrate the asynchronous propagation and intra\hyp island synchronization of the population using the example of the blue individual {\small $\mathrm{ind}_{g3}^{i1}$}. 
    This individual is bred by worker 1 in generation 3 by applying the propagator (yellow) to the worker's current population. 
    After evaluating {\small $\mathrm{ind}_{g3}^{i1}$}, worker 1 sends it to all workers on its island and appends it to its population. 
    As no evaluated individuals dispatched by worker 2 await to be received, worker 1 proceeds with breeding. 
    Worker 2 receives the blue {\small $\mathrm{ind}_{g3}^{i1}$} only after finishing the evaluation of the red {\small $\mathrm{ind}_{g2}^{i1}$}. 
    It then appends both to its population and breeds a new individual for generation 3. 
    }
    \label{fig:async_propagation}
\end{figure}

\fancyname's basic mechanism is that of Darwinian evolution, i.e., beneficial traits are selected, recombined, and mutated to breed more fit individuals (see \Cref{alg:basicGA}). 
On a higher level, \fancyname employs an IM, which combines independent evolution of self\hyp contained subpopulations with intermittent exchange of selected individuals~\cite{sudholt2015pea}. 
To coordinate the search globally, each island occasionally delegates migrants to be included in the target islands' populations. 
Islands communicate genetic information competitively, thus increasing diversity among the subpopulations compared to panmictic models~\cite{cantupaz2000efficient}.  
For synchronous IMs, this exchange occurs simultaneously after fixed intervals, with no computation happening in that time. 
The following hyperparameters characterize IMs:
\begin{itemize}
    \item \textbf{Island number and subpopulation sizes}
    \item \textbf{Migration (pollination) probability}
    \item \textbf{Number of migrants (pollinators):} How many individuals migrate from the source population at a time. 
    \item \textbf{Migration (pollination) topology:} Directed graph of migration (pollination) paths between islands. 
    \item \textbf{Emigration policy:} How to select emigrants (e.g., random or best) and whether to remove them from the source population (actual migration) or not (pollination).
    \item \textbf{Immigration policy:} How to insert immigrants into the target population, i.e., either add them (migration) or replace existing individuals (pollination, e.g., random or worst). 
\end{itemize}
%
\begin{algorithm}[!hp]
\small
\SetAlgoLined
\DontPrintSemicolon 
\SetKwBlock{DoParallel}{for}{end}
\SetKwInOut{Input}{Input}
\Input{Search\hyp space limits; hyperparameters \textit{n\_islands}, island sizes $P_i$ ($i=1,\dots,$ \textit{n\_islands}), number of iterations \textit{generations}, evolutionary operators (including \textit{selection\_policy},
\textit{crossover\_probability}, \textit{mutation\_probability} etc.), \textit{pollination\_probability}, \textit{pollination\_topology}, \textit{emigration\_policy}, \textit{immigration\_policy}.}
\SetKwBlock{DoParallel}{for}{end}
Configure \textit{n\_islands} islands with $P_i$ workers each. 
Each worker evaluates one individual at a time and maintains its own population list \textit{pop} of evaluated and migrated individuals on the island.\;
\tcc{START OPTIMIZATION.}
\DoParallel(each worker \textbf{do in parallel}){
\While(\tcp*[h]{Loop over generations.}){generation $\leq$ generations}{
    Breed and evaluate individual. 
    Append it to \textit{pop}. 
    Send it to other workers on island to synchronize their populations lists: \texttt{evaluate\_individual()}\tcp*[r]{BREED AND EVALUATE}
    Check for and possibly receive individuals bred and evaluated by other workers on island. 
    Append them to \textit{pop}: \texttt{receive\_intra\_isle\_individuals()} \tcp*[r]{SYNCHRONIZE}
    \If(\tcp*[f]{EMIGRATE}){random $\leq$ pollination\_probability}{
        Choose pollinators from currently active individuals on island according to \textit{emigration\_policy}. 
        Send copies of pollinator(s) to workers of target islands according to \textit{pollination\_topology}: 
        \texttt{send\_emigrants()}
        \;}
    Check for and possibly receive pollinators sent by workers from other islands. 
    Add them to \textit{pop}. 
    Determine individuals to be replaced by incoming pollinators according to \textit{immigration\_policy}. 
    Send individuals to be replaced to other workers on island for deactivation: \texttt{receive\_immigrants()}\tcp*[r]{IMMIGRATE}
    Check for and possibly receive individuals replaced by pollinators on other workers on island. 
    Try to deactivate them in \textit{pop}. 
    If an individual to be deactivated is not yet in \textit{pop}, append it to history list \textit{replaced} and try again in the next generation: \texttt{deactivate\_replaced\_individuals()}\tcp*[r]{SYNCHRONIZE}
    Go to next generation: \texttt{generation += 1}
}
\tcc{OPTIMIZATION DONE: FINAL SYNCHRONIZATION}
Wait for all other workers to finish: \texttt{MPI.COMM\_WORLD.barrier()}\;
Final check for incoming messages so all workers hold complete population.\;
Probe individuals evaluated by other workers on island: \texttt{receive\_intra\_isle\_individuals()}\;
Probe for incoming pollinators immigrating from other islands: \texttt{receive\_immigrants()}\;
Probe for individuals replaced by other workers on island to be deactivated:  \texttt{deactivate\_replaced\_individuals()}\;
}
\KwResult{$n$ individuals with smallest OF values.}
\caption{\textbf{\fancyname with pollination.}}
 \label{alg:propulate_pollination}
\end{algorithm}
\fancyname's functional principle is outlined in \Cref{alg:propulate_pollination}. 
We consider multiple PEs (or workers) partitioned into islands. 
Each worker processes one individual at a time and maintains a population to track evaluated and migrated individuals on its island. 
To mitigate the computational overhead of synchronized OF evaluations, \fancyname leverages asynchronous propagation of continuous populations with interwoven, worker\hyp specific generations (see \Cref{fig:async_propagation}). 
In each iteration, each worker breeds and evaluates an individual which is added to its population list. 
It then sends the individual with its evaluation result to all workers on the same island and, in return, receives evaluated individuals dispatched by them for a mutual update of their population lists. 
To avoid explicit synchronization points, the independently operating workers use asynchronous point\hyp to\hyp point communication via MPI to share their results. 
Each one dispatches its result immediately after finishing an evaluation. 
Directly afterwards, it non\hyp blockingly checks for incoming messages from workers of its own island awaiting to be received. 
In the next iteration, it breeds a new individual by applying the evolutionary operators to its continuous population list of all evaluated individuals from any generation on the island. 
The workers thus proceed asynchronously without idle times despite the individuals' varying computational costs. 

After the mutual update, asynchronous migration or pollination between islands happens on a per\hyp worker basis with a certain probability. 
Each worker selects a number of emigrants from its current population. 
For actual migration\footnote{See \href{https://github.com/Helmholtz-AI-Energy/propulate/tree/master/supplementary}{github.com/Helmholtz-AI-Energy/propulate/tree/master/supplementary} for pseudocode with migration and explanatory figure.}, an individual can only exist actively on one island. 
A worker thus may only choose eligible emigrants from an exclusive subset of the island's population to avoid overlapping selections by other workers. 
It then dispatches the emigrants to the target islands' workers as specified in the migration topology. 
Finally, it sends them to all workers on its island for island\hyp wide deactivation of emigrated individuals before deactivating them in its own population. 

In the next step, the worker probes for and, if applicable, receives immigrants from other islands. 
It then checks for individuals emigrated by other workers of its island and tries to deactivate them in its population. 
Due to the asynchronicity, individuals might be designated to be deactivated before arriving in the population. 
\fancyname continuously corrects these synchronization artefacts during the optimization. 

\begin{figure}[tb]
    \centering
    \includegraphics[width=1.0\linewidth]{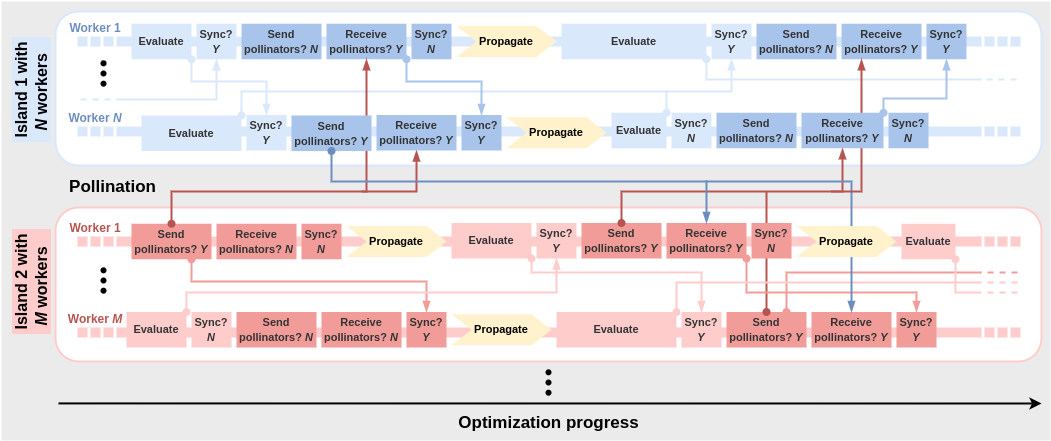}
    \setlength{\belowcaptionskip}{-10pt} 
    \caption{\textbf{Asynchronous pollination.} 
    Consider two islands with $N$ (blue) and $M$ (red) workers, respectively. 
    We illustrate pollination (dark colors) by tracing worker $N$ on island 1. 
    After evaluation and mutual intra\hyp island updates (light blue, see \Cref{fig:async_propagation}), this worker performs pollination: it sends copies of the chosen pollinators to all workers of each target island, here island 2. 
    The target island's workers receive the pollinators asynchronously (dark blue arrows). 
    For proper accounting of the populations, worker 1 on island 2, selects the individual to be replaced and informs all workers on its island accordingly (middle red arrow). 
    Afterwards, worker $N$ receives incoming pollinators from island 2 to be included into its population. 
    It then probes for individuals that have been replaced by other workers on its island, here worker 1, in the meantime and need to be deactivated. 
    After these pollination\hyp related intra\hyp island population updates, it breeds the next generation. 
    As pollination does not occur in this generation, it directly receives pollinators from island 2. 
    This time, worker $N$ chooses the individual to be replaced. 
    }
    \label{fig:async_pollination}
\end{figure}

For pollination (\Cref{fig:async_pollination}), identical copies of individuals can exist on multiple islands. 
Workers thus can choose emigrating pollinators from any active individuals in their current populations and do not deactivate them upon emigration. 
To control the population growth, pollinators replace active individuals in the target population according to the immigration policy. 
For proper accounting of the population, one random worker of the target island selects the individual to be replaced and informs the other workers accordingly. 
Individuals to be deactivated that are not yet in the population are cached to be replaced in the next iteration. 
This process is repeated until each worker has evaluated a set number of generations. 
Finally, the population is synchronized among workers and the best individuals are returned. 

\fancyname uses so\hyp called propagators to breed child individuals from an existing collection of parent individuals. 
It implements various standard genetic operators, including uniform, best, and worst selection, random initialization, stochastic and conditional propagators, point and interval mutation, and several forms of crossover. 
In addition, \fancyname provides a default propagator:
Having selected two random parents from the breeding pool consisting of a set number of the currently most fit individuals, uniform crossover and point mutation are performed each with a specified probability. 
Afterwards, interval mutation is performed. 
To prevent premature trapping in a local optimum, a randomly initialized individual is added with a specified probability instead of one bred from the current population.
\section{Experimental Evaluation}
\label{sec:experimental-evaluation}
We evaluate \fancyname on various benchmark functions (see \Cref{subsec:function-optimization}) and an HPO use case in remote sensing classification (see \Cref{subsec:nn-optimization}) which provides a real world application.
We compare our results against \texttt{Optuna}, since it is the most widely used HPO software.

\subsection{Experimental Environment}
\label{subsec:experimental-environment}
We ran the experiments on the distributed\hyp memory, parallel hybrid supercomputer \textit{Hochleistungsrechner Karlsruhe} (HoreKa) at the Steinbuch Centre for Com- puting, Karlsruhe Institute of Technology. 
Each of its 769 compute nodes is equipped with two 38\hyp core Intel Xeon Platinum 8368 processors at 2.4 GHz base and 3.4 GHz maximum turbo frequency, 256 GB (standard) or 512 GB (high\hyp memory and accelerator) local memory, a local 960 GB NVMe SSD disk, and two network adapters.
167 of the nodes are accelerator nodes each equipped with four NVIDIA A100\hyp 40 GPUs with 40 GB memory connected via NVLink. 
Inter\hyp node communication uses a low\hyp latency, non\hyp blocking NVIDIA Mellanox InfiniBand 4X HDR interconnect with 200 Gbit/s per port. 
A Lenovo Xclarity controller measures full node energy consumption, excluding file systems, networking, and cooling. 
The operating system is Red Hat Enterprise Linux 8.2. 

\subsection{Benchmark Functions}
\label{subsec:benchmark-function}
\begin{table}[t]
\small
\centering
\renewcommand{\arraystretch}{1.25}
\caption{\textbf{Benchmark functions}.\vspace{0.2cm}}
\begin{tabular}{llll}
\hline
Name & Function & Limits & Global minimum\\
\hline
Sphere & \(f_1=x_1^2+x_2^2\) & \(\pm5.12\) & \(f\left(0,0\right)=0\)\\
Rosenbrock & \(f_2 = 100\left(x_1^2-x_2\right)^2+\left(1-x_1\right)^2\)& \(\pm 2.048\) & \(f\left(1,1\right)=0\)\\
Step & \(f_3=\sum_{i=1}^5 \mathrm{int}\left(x_i\right)\) & \(\pm5.12\) & \(f\left(x_i\leq-5\right)=-25\)\\
Quartic & \(f_4=\sum_{i=1}^{30}\left(ix_i^4 + \mathcal{N}_i\left(0,1\right)\right)\) & \(\pm 1.28\) & \(f\left(0,...,0\right)=\sum_i\mathcal{N}_i\) \\
Rastrigin & \(f_5=200+\sum_{i=1}^{20}x_i^2-10\cos\left(2\pi x_i\right) \) & \(\pm 5.12\) & \(f\left(0,...,0\right)=0\)\\
Griewank & \(f_6=1+\frac{1}{4000}\sum_{i=1}^{10}x_i^2-\prod_{i=1}^{10}\cos\frac{x_i}{\sqrt{i}}\) & \(\pm 600\) & \(f\left(0,...,0\right)=0\)\\
Schwefel & \(f_7=10V-\sum_{i=1}^{10}x_i\sin\sqrt{\left|x_i\right|}\) & \(\pm 500\) & \(f\left(x_1^*,...,x_{10}^*\right)=0\text{,}\)\\
& \(\text{with } V=418.982887\) & & $x_i^*=420.968746$ \\
Bi\hyp sphere & \(f_8= \min\left(\sum_{i=1}^{30} \left(x_i-\mu_1\right)^2, \right. \) & \(\pm 5.12\) & \(f\left(\mu_1,...,\mu_1\right)=0\)\\
& \(\left. 30+s\cdot \sum_{i=1}^{30} \left(x_i-\mu_2\right)^2\right)\text{ with}\)& & \\
& \(\mu_1=2.5\text{, }\mu_2=-\left(s^{-1}\left(\mu_1^2-1\right)\right)^{1/2}\text{, }\)& & \\
& \(s=1-\left(2\sqrt{50}-8.2\right)^{-1/2}\) & & \\
Bi\hyp Rastrigin & \(f_9= f_8+10\sum_{i=1}^{30} 1-\cos2\pi\left(x_i-\mu_1\right)\quad\) & \( \pm 5.12\) & \(f\left(\mu_1,...,\mu_1\right)=0\)\\
\hline
\end{tabular}
\label{tab:math-fns}
\end{table}
Benchmark functions are used to evaluate optimizers in terms of convergence, accuracy, and robustness. 
The informative value of such studies is limited by how well we understand the characteristics making real\hyp life optimization problems difficult and our ability to embed these features into benchmark functions~\cite{lunacek2019impact}. 
We use \fancyname to optimize a variety of traditional and recent benchmark functions emulating situations optimizers have to cope with in different kinds of problems (see \Cref{tab:math-fns}). 

\begin{itemize}
\item \textbf{Sphere} is smooth, unimodal, strongly convex, symmetric, and thus simple.

\item \textbf{Rosenbrock} has a narrow minimum inside a parabola\hyp shaped valley. 

\item \textbf{Step} represents the problem of flat surfaces. 
Plateaus pose obstacles to optimizers as they lack information about which direction is favorable. 

\item \textbf{Quartic} is a unimodal function padded with Gaussian noise. As it never returns the same value on the same point, algorithms that do not perform well on this test function will do poorly on noisy data. 

\item \textbf{Rastrigin} is non\hyp linear and highly multimodal.
Its surface is determined by two external variables, controlling the modulation's amplitude and frequency. 
The local minima are located at a rectangular grid with size 1. 
Their functional values increase with the distance to the global minimum. 

\item \textbf{Griewank}'s product creates sub\hyp populations strongly codependent to parallel GAs, while the summation produces a parabola. 
Its local optima lie above parabola level but decrease with increasing dimensions, i.e., the larger the search range, the flatter the function. 

\item \textbf{Schwefel} has a second\hyp best minimum far away from the global optimum. 

\item \textbf{Lunacek's bi\hyp sphere}'s~\cite{lunacek2019impact} landscape structure is the minimum of two quadratic functions, each creating a single funnel in the search space.
The spheres are placed along the positive search\hyp space diagonal, with the optimal and sub\hyp optimal sphere in the middle of the positive and negative quadrant, respectively. 
Their distance and the barrier's height increase with dimensionality, creating a globally non\hyp separable underlying surface. 

\item \textbf{Lunacek's bi\hyp Rastrigin~\cite{lunacek2019impact}} is a double\hyp funnel version of Rastrigin. 
This function isolates global structure as the main difference impacting problem difficulty on a well understood test case. 
\end{itemize}

\subsection{Meta-Optimizing the Optimizer}
\fancyname itself has HPs influencing its optimization behavior, accuracy, and robustness. 
To explore their effect systematically and give transparent recommendations for default values, we conducted a grid search across the six most prominent HPs.
The search space is shown in \Cref{tab:grid-search-params}.
We ran the grid search five times for the quartic, Rastrigin, and bi\hyp Rastrigin benchmark functions (see \Cref{tab:math-fns} and \Cref{subsec:function-optimization}), each with a different seed consistently used over all points within a search. 
All three functions have their global minimum at zero. 
They were chosen for their high\hyp dimensional parameter spaces (30, 20, and 30, respectively) and different levels of difficulty to optimize.
\begin{table}[tb]
\caption{\textbf{Grid search parameters.} 
All experiments use 144 CPUs equally distributed between two nodes. 
Random\hyp initialization probability refers to the chance that a new individual is generated entirely randomly.\vspace{0.2cm}}
\label{tab:grid-search-params}
\centering
\begin{tabular}{@{}l@{\hspace{2em}}c@{\hspace{2em}}c@{\hspace{2em}}c@{\hspace{2em}}c@{\hspace{2em}}c@{}}
\toprule
Number of islands                 & 2          & 4           & 8       & 16          & 32          \\
Island population size            & 72         & 36          & 18      & 9           & 4           \\
Migration (pollination) probability             & 0.1        & 0.3         & 0.5     & 0.7         & 0.9         \\
Pollination                       & \multicolumn{2}{c}{True} & \multicolumn{3}{c}{False}           \\
Crossover probability             & 0.1        & 0.325       & 0.55    & \multicolumn{2}{c}{0.775} \\
Point-mutation probability        & 0.1        & 0.325       & 0.55    & \multicolumn{2}{c}{0.775} \\
Random-initialization probability & 0.1        & 0.325       & 0.55    & \multicolumn{2}{c}{0.775} \\
\bottomrule
\end{tabular}
\end{table}
For quartic, \fancyname found a minimum below $0.01\pm0.005$ for 80.12\,\% of all points across the five grid searches.
This increases to 94.94\,\% for minima found within $0.1\pm0.05$ of the global minimum.
In comparison, the tolerances have to be relaxed considerably for the more complex Rastrigin and bi\hyp Rastrigin.
While only 18.57\,\% of all grid points had a function value less than $1.0\pm0.5$ for Rastrigin, only a single point resulted in an average value of less than 10 for bi\hyp Rastrigin.
Although the average value of bi\hyp Rastrigin was only less than 10 once, we found the minimum across each of the five searches to be less than 1.0 for 3.31\,\% of the grid points. 

Considering grid points with at least one result smaller than 1.0, 86.61\,\% used either 16 or 36 islands, while the remainder used eight.
As \fancyname initializes different islands at different positions in the search space, the chance that one of them is at a very beneficial position increases with the number of islands.
This is further confirmed by a migration probability of 0.7 or 0.9 for 61.41\,\% of these points.
If one of the islands is well\hyp initialized, it thus will quickly notify others.

With every best grid point using pollination, we clearly find pollination to be favorable over real migration.
To determine the other HPs, we compute the averages of the results for the top ten grid points across all three functions.
The top ten were determined by grouping over the lowest average and standard deviation of the function values, sorting by the averages, and sorting by the standard deviations.
This method reduces the chances of a single run simply benefiting from an advantageous starting seed.
Average crossover, point\hyp mutation, and random\hyp initialization probabilities are $0.655 \pm 0.056$, $0.363 \pm 0.133$, and $0.423 \pm 0.135$, respectively.
The average number of islands was $28.800 \pm 6.009$ which equates to an island population of $5.00 \pm 1.043$.
The average migration probability was $0.527 \pm 0.150$. 
These values provide a reasonable starting point towards choosing default HPs for \fancyname (see \Cref{tab:function-hps}).
As the grid searches only considered functions with independent parameters, we assume a relatively high random\hyp initialization probability to be useful due to the benefits of random search~\cite{bergstra2012random}.
On this account, we chose to reduce the default random\hyp initialization probability to 0.2.
As the migration probability might also be lowered artificially by this phenomenon, we set its default to 0.7. 
The default probabilities for crossover and point\hyp mutation were chosen as 0.7 and 0.4, respectively.
The island size was set at four individuals.
This is a practical choice as our test system has four accelerators per node and the number of CPUs per node is a multiple of four. 

\subsection{Benchmark Function Optimization}
\label{subsec:function-optimization}
\begin{figure}[tb]
    \centering
    \includegraphics[width=0.85\linewidth]{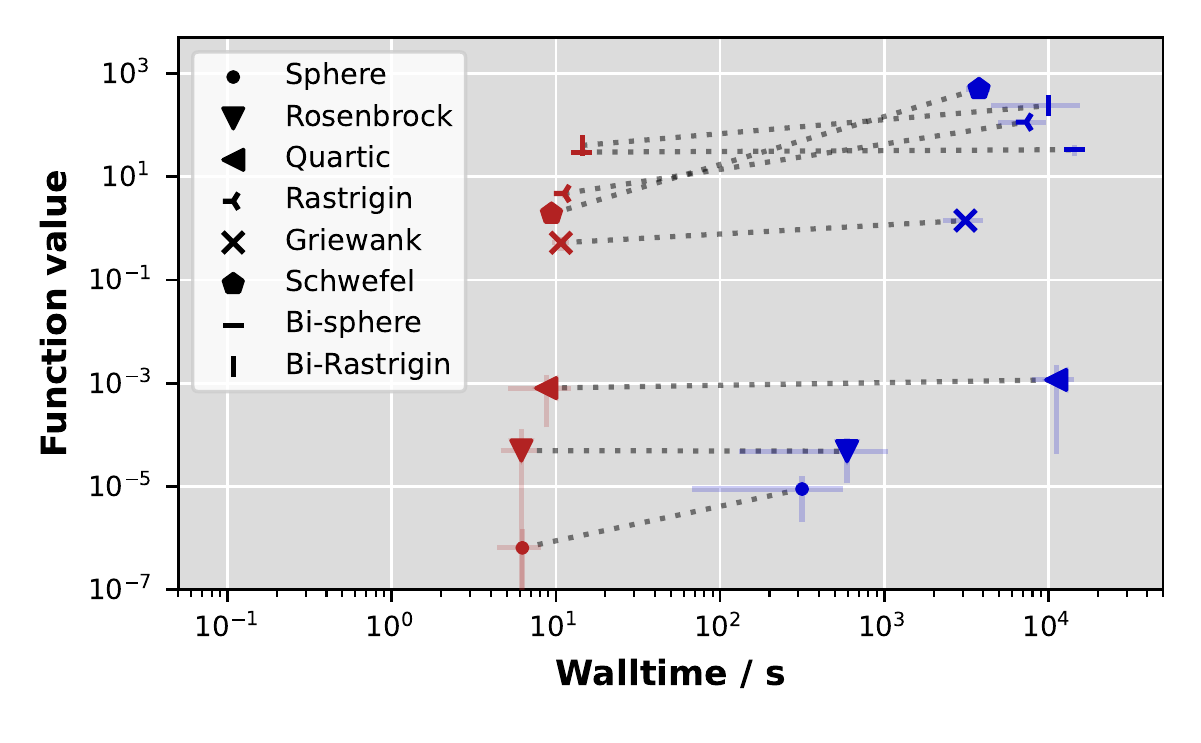}
    \setlength{\belowcaptionskip}{-10pt} 
    \caption{\textbf{Benchmark function minimization accuracy over walltime.} 
    Lowest function values found by \fancyname (red) and \texttt{Optuna} (blue) versus walltimes to reach them, each averaged over ten runs. 
    Step is not shown since both optimizers achieve a perfect value of $-25$ within 0.6 s  and 278.2 s, respectively. 
    }
    \label{fig:function_benchmark}
\end{figure}

For each function, we ran each ten equivalent \fancyname and \texttt{Optuna} optimizations, using the same compute resources, degree of parallelization, and number of evaluations. 
\begin{table}[tb]
\footnotesize
\centering
\caption{\textbf{\fancyname HPs for benchmark function minimization.} \vspace{0.2cm}}
\label{tab:function-hps}
\begin{tabular}{@{}ll@{}}
\toprule
Number of islands                 & 38                   \\
Island population size                & 4                    \\
Pollination probability             & 0.7                  \\
Crossover probability             & 0.7                  \\
Point-mutation probability        & 0.4                  \\
Sigma factor                      & 0.05                 \\
Random-initialization probability$\qquad$ & 0.2                  \\
Generations per worker             & 256                \\
Selection policy                  & Best                 \\
Pollination topology              & Fully connected      \\
Number of migrants                & 1                    \\
Emigration policy                 & Best \\
Immigration policy                & Worst \\
\bottomrule
\end{tabular}
\end{table}
\Cref{fig:function_benchmark} shows the optimization accuracy over walltime comparing \fancyname with default parameters determined from our grid search (see \Cref{tab:function-hps}) to \texttt{Optuna}'s default optimizer. 
In terms of accuracy, \fancyname and \texttt{Optuna} are comparable in most experiments. 
For many functions, e.g. Schwefel, bi\hyp Rastrigin, and Rastrigin, \fancyname even achieves a better OF value. 
In terms of walltime, \fancyname is consistently at least one order of magnitude faster. 
This is due to \fancyname's MPI\hyp based communication over the fast network, whereas \texttt{Optuna} uses relational databases with SQL and is limited by the slow file system. 
Since the functions are cheap to evaluate, optimization and communication dominate the walltime. 
In particular for problems where evaluations are cheap compared to the search itself, we find that \texttt{Optuna}'s computational efficiency suffers massively from the frequent file locking inherent to its parallelization strategy, reducing its usability for large\hyp scale HPC applications. 

\begin{figure}[htb]
    \centering
    \includegraphics[width=0.85\linewidth]{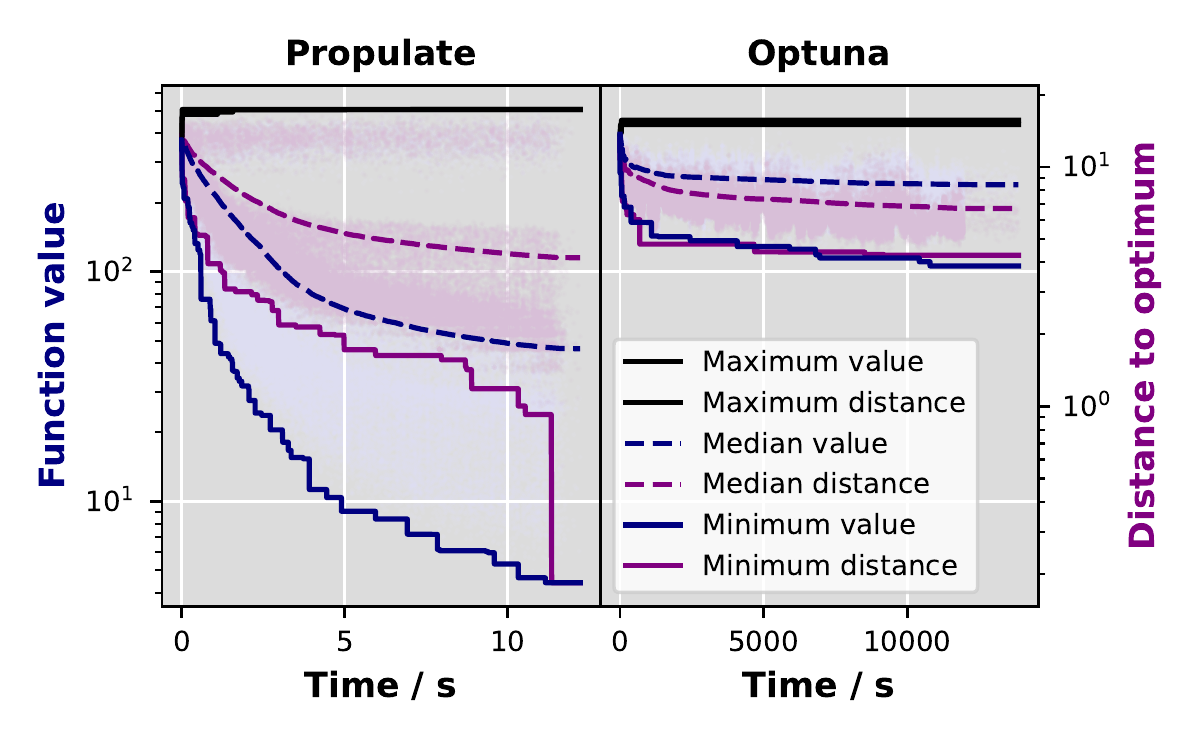}
    \caption{\textbf{Evolution of the population over walltime for the Rastrigin function.} 
    \fancyname (left) versus \texttt{Optuna} (right).
    OF values (blue) use the left\hyp hand scale, distances to the global optimum (purple) use the right\hyp hand scale.
    Pastel dots show each individual's OF value/distance.
    Solid (dashed) lines show the minimum (median) value and distance achieved so far.
    Maximum value (distance) are shown in black.
    Both optimizers perform 38\,912 evaluations.
    Note the difference on the time axis.
    }
    \label{fig:function_benchmark_evolution}
\end{figure}

In addition, we inspected the evolution of the population over walltime for both \fancyname and \texttt{Optuna}. 
An example for minimizing the Rastrigin function is shown in \Cref{fig:function_benchmark_evolution}. 
\fancyname is roughly three orders of magnitude faster and makes significantly greater progress in terms of both OF values and distance to the global optimum.
Due to this drastic difference in runtime, we measure only 46.27 Wh for \fancyname compared to \texttt{Optuna}'s 2646.29 Wh.
\subsection{HP Optimization for Remote Sensing Classification}
\label{subsec:nn-optimization}
BigEarthNet~\cite{sumbul2020bigearthnet} is a Sentinel\hyp 2 multispectral image dataset in remote sensing. 
It comprises 590\,326 image patches each of which is assigned one or more of the 19 available CORINE Land Cover map labels~\cite{bossard2000corine,sumbul2020bigearthnet}. 
Multiple computer vision networks for BigEarthNet classification have been trained~\cite{sumbul2020bigearthnet}, with ResNet\hyp 50~\cite{he2016deep} being the most accurate. 
While a previous \fancyname version was used to optimize a set of HPs and the architecture for this use case~\cite{coquelin2021evolutionary}, a more versatile and efficient parallelization strategy in the current version makes it worthwhile to revisit this application.
Analogously to \cite{coquelin2021evolutionary}, we consider different optimizers, learning rate (LR) schedulers, activation functions, loss functions, number of filters in each convolutional block, and activation order~\cite{he2016identity}. 
The search space is shown in \Cref{tab:nn-params}. 
Optimizer parameters, LR functions, and LR warmup are included as well. 
We only consider SGD\hyp based optimizers as they share common parameters and thus exclude Adam\hyp like optimizers from the search.
We theorize that including Adam led to the difficulties seen previously~\cite{coquelin2021evolutionary}. 
The training is exited if the validation loss has not been increasing for ten epochs. 
We prepared the data analogously to \cite{coquelin2021evolutionary}. 
The network is implemented in TensorFlow~\cite{abadi2016tensorflow}.

For both \fancyname and \texttt{Optuna}, we ran each three searches over 24 h on 32 GPUs. 
We use $1-F_1^{\text{val}}$ with the validation $F_1$ score as the OF to be minimized. 
On average, \texttt{Optuna} achieves its best OF value of (0.39 $\pm$ 0.01) h within (7.05 $\pm$ 3.14) h.
\fancyname beats \texttt{Optuna}'s average best after (5.30 $\pm$2.41) h and achieves its best OF value of $\left(0.36 \pm 0.00\right)$ within (13.89 $\pm$ 5.15) h.
\begin{table}[htb]
\renewcommand{\arraystretch}{1.05}
\caption{\textbf{HP search space of ResNet\hyp 50 for BigEarthNet classification.\vspace{0.2cm}}}
\label{tab:nn-params}
\begin{tabular}{@{}lllll@{}}
\toprule
\small
Optimizers & \multicolumn{2}{c}{Optimizer parameters}                 & \multicolumn{2}{c}{LR warmup parameters}                                              \\ \cmidrule(r){1-1} \cmidrule(l){2-3} \cmidrule(l){4-5} 
Adagrad        & Initial accum. value & $\left[10^{-4},0.5\right]$        & LR warmup steps                         & $\left[10^0, 10^4\right]$                         \\
SGD            & Clipnorm                  & $\left[-1,-1000\right]$         & Initial LR                   & $\left[10^{-5},10^{-1}\right]$                        \\
Adadelta       & Clipvalue                 & $\left[-1,1000\right]$         & Decay steps                             & $\left[10^2,10^5\right]$                         \\
RMSprop        & Use EMA                   & Boolean           & LR warmup power                         & $\left[10^{-1},10^1\right]$                          \\
               & EMA momentum              & $\left[0.5,1.0\right]$         &                                         &                                   \\
               & EMA overwrite             & $\left[1, 10^3\right]$          &                                         &                                   \\
               & Momentum                  & $\left[0.0,1.0\right]$         &                                         &                                   \\
               & Nesterov                  & Boolean           &                                         &                                   \\
               & Rho                       & $\left[0.8,0.99999\right]$     &                                         &                                   \\
               & Epsilon                   & $\left[10^{-9},10^{-4}\right]$       &                                         &                                   \\ \midrule
\multicolumn{3}{c}{Loss functions}                              & \multicolumn{2}{c}{LR parameters}                                                      \\ \cmidrule(r){1-3} \cmidrule(r){4-5}
Binary CE      & Categorical CE            & Categorical hinge & \multicolumn{1}{l}{Decay rate}          & \multicolumn{1}{l}{$\left[0.8,0.9999\right]$}  \\
Hinge          & KL divergence              & Squared hinge     & \multicolumn{1}{l}{Staircase inverse}           & \multicolumn{1}{l}{Boolean}       \\
               &                           &                   & \multicolumn{1}{l}{time decay}    & \multicolumn{1}{l}{}              \\ \cmidrule(r){1-3}
\multicolumn{3}{c}{Activation functions}                              & \multicolumn{1}{l}{Decay rate}          & \multicolumn{1}{l}{$\left[0.1,0.9\right]$}     \\ \cmidrule(r){1-3}
ELU            & ReLU                      & Softplus          & \multicolumn{1}{l}{Staircase poly-}           & \multicolumn{1}{l}{Boolean}       \\
Exponential    & SELU                      & Softsign          & \multicolumn{1}{l}{nomial decay}     & \multicolumn{1}{l}{}              \\
Hard sigmoid  & Sigmoid                   & Swish             & \multicolumn{1}{l}{End LR} & \multicolumn{1}{l}{$\left[10^{-4},10^{-2}\right]$} \\
Linear         & Softmax                   & Tanh              & \multicolumn{1}{l}{Power}               & \multicolumn{1}{l}{$\left[0.5,2.5\right]$}    
\end{tabular}
\end{table}

\subsection{Scaling}
\begin{figure}[htb]
    \centering
    \includegraphics[width=.85\textwidth]{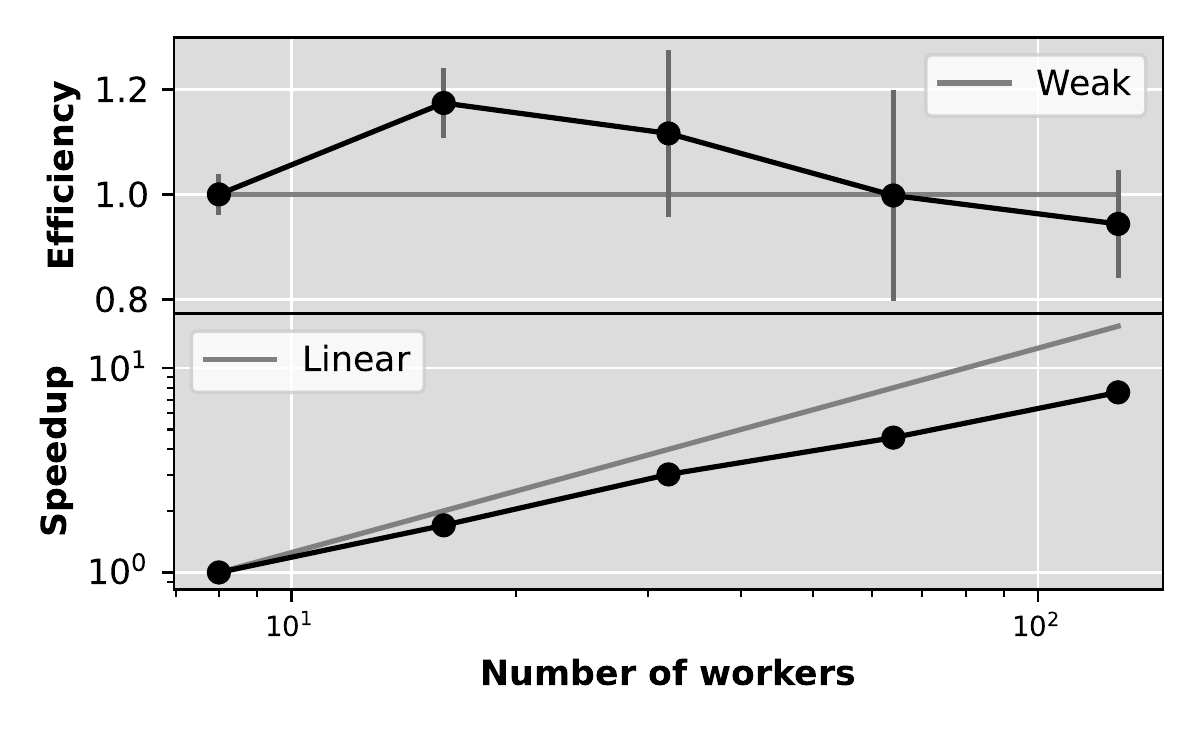}
    \caption{\textbf{Scaling with respect to a baseline of eight workers.} 
    Weak efficiency (top) and strong linear speedup (bottom). 
    Use case and search space are described in \Cref{subsec:nn-optimization}. 
    Weak\hyp scaling problem size is varied via the number of OF evaluations. 
    Results are averaged over three runs.}
    \label{fig:scaling}
\end{figure}
Finally, we explore \fancyname's scaling behavior for the use case presented in \Cref{subsec:nn-optimization}.
\Cref{fig:scaling} shows our results for weak and strong linear scaling. 
Our baseline configuration used two nodes.
Since each node has four GPUs, we calculate speedup and efficiency with respect to eight workers.
For strong scaling, we fix the total number of evaluations at $512$ and increase the number of workers, i.e., GPUs.
We average over three runs with different seeds and keep four workers per island while increasing the number of islands.
Speedup increases up to 128 workers, where we reach approximately half the optimal value.
This is an expected decline since each worker only processes few individuals, so the variance in evaluation times leads to larger idle times of the faster workers before the final population synchronization at the end.
Additionally, as the number of workers approaches the total number of evaluations, the randomly initialized evolutionary search in turn approaches a random search.
This means that the search performance is likely to be worse than what the pure compute performance might suggest.
It is still possible to apply \fancyname on these scales, but the other search parameters have to be adjusted accordingly as shown in the weak scaling plot \Cref{fig:scaling} top.
Weak efficiency only drops to 95\, \% at our largest configuration of 128 workers
The early super\hyp scalar behavior is likely due to the non\hyp sequential baseline.
\section{Conclusion}
\label{sec:concolusion}
We presented \fancyname, our HPC\hyp adapted, asynchronous genetic optimization algorithm and software.
Our experimental evaluation shows that the fully asynchronous evaluation, propagation, and migration enable a highly efficient and parallelizable genetic optimization.
Harder to quantify than performance but very important is ease of use.
Especially for HPC applications at scale, some parallelization and distribution models are more suited than others.
A purely MPI\hyp based implementation as in \fancyname is not only extremely efficient for highly parallel and communication\hyp intensive algorithms but also easy to set up and maintain, since the required infrastructure is commonly available on HPC systems. 
This is not the case for any of the other tools investigated, except for the not publicly available \texttt{MENNDL}.
This also facilitates a tighter coupling of individuals during the optimization, which enables a more efficient evaluation of candidates and in particular early stopping informed by previously evaluated individuals in the NAS case.
\fancyname was already successfully applied to HPO for various ML models on different HPC machines~\cite{coquelin2021evolutionary,funk2022prediction}. 
Another avenue for future work is including variable\hyp length gene descriptions.
Mutually exclusive genes of different lengths, such as the parameter sets for Adam- and SGD\hyp like optimizers in our NAS use case, can thus be explored efficiently.
While this is already possible, it requires an inconvenient workaround of including inactive genes and adapting the propagators to manually prevent the evaluation of many individuals differing only in inactive genes.

\section*{Acknowledgments}
\label{sec:acknowledgments}
This work is supported by the Helmholtz AI platform grant and the Helmholtz Association Initiative and Networking Fund on the HAICORE@KIT partition.
\bibliographystyle{splncs04}
\bibliography{references}

\begin{thebibliography}{10}
\providecommand{\url}[1]{\texttt{#1}}
\providecommand{\urlprefix}{URL }
\providecommand{\doi}[1]{https://doi.org/#1}

\bibitem{abadi2016tensorflow}
Abadi, M., Barham, P., Chen, J., Chen, Z., Davis, A., Dean, J., Devin, M.,
  Ghemawat, S., Irving, G., Isard, M., et~al.: {TensorFlow: A System for
  Large-Scale Machine Learning}. In: {12th USENIX Symposium on Operating
  Systems Design and Implementation (OSDI 16)}. pp. 265--283 (2016)

\bibitem{akiba2019optuna}
Akiba, T., Sano, S., Yanase, T., Ohta, T., Koyama, M.: {Optuna: A
  Next-generation Hyperparameter Optimization Framework}. In: Proceedings of
  the 25th ACM SIGKDD International Conference on Knowledge Discovery \& Data
  Mining. pp. 2623--2631 (2019). \doi{10.1145/3292500.3330701}

\bibitem{alba2002parallelism}
Alba, E., Tomassini, M.: {Parallelism and Evolutionary Algorithms}. IEEE
  Transactions on Evolutionary Computation  \textbf{6}(5),  443--462 (2002).
  \doi{10.1109/TEVC.2002.800880}

\bibitem{alba1999ASO}
Alba, E., Troya, J.M.: {A Survey of Parallel Distributed Genetic Algorithms}.
  Complexity  \textbf{4}(4),  31--52 (1999)

\bibitem{gpyopt2016}
Authors, T.G.: {GPyOpt: A Bayesian Optimization Framework in Python}.
  \url{http://github.com/SheffieldML/GPyOpt} (2016)

\bibitem{bergstra2012random}
Bergstra, J., Bengio, Y.: {Random Search for Hyper-Parameter Optimization}.
  Journal of Machine Learning Research  \textbf{13}(10),  281--305 (2012),
  \url{http://jmlr.org/papers/v13/bergstra12a.html}

\bibitem{bergstra2013making}
Bergstra, J., Yamins, D., Cox, D.: {Making a Science of Model Search:
  Hyperparameter Optimization in Hundreds of Dimensions for Vision
  Architectures}. In: International Conference on Machine Learning. pp.
  115--123. PMLR (2013), \url{http://proceedings.mlr.press/v28/bergstra13.pdf}

\bibitem{bianchi2009survey}
Bianchi, L., Dorigo, M., Gambardella, L.M., Gutjahr, W.J.: A survey on
  metaheuristics for stochastic combinatorial optimization. Natural Computing
  \textbf{8}(2),  239--287 (2009). \doi{10.1007/s11047-008-9098-4}

\bibitem{blum2003metaheuristics}
Blum, C., Roli, A.: {Metaheuristics in Combinatorial Optimization: Overview and
  Conceptual Comparison}. ACM Computing Surveys (CSUR)  \textbf{35}(3),
  268--308 (2003). \doi{10.1145/937503.937505}

\bibitem{bossard2000corine}
Bossard, M., Feranec, J., Otahel, J., et~al.: {CORINE land cover technical
  guide: Addendum 2000}, vol.~40. European Environment Agency Copenhagen (2000)

\bibitem{cantupaz2000efficient}
Cant{\'u}-Paz, E.: {Efficient and Accurate Parallel Genetic Algorithms},
  vol.~1. Springer Science \& Business Media (2000).
  \doi{10.1007/978-1-4615-4369-5}

\bibitem{cantupaz1998survey}
Cant{\'u}-Paz, E., et~al.: {A Survey of Parallel Genetic Algorithms}.
  Calculateurs paralleles, reseaux et systems repartis  \textbf{10}(2),
  141--171 (1998)

\bibitem{coquelin2021evolutionary}
Coquelin, D., Sedona, R., Riedel, M., G{\"o}tz, M.: {Evolutionary Optimization
  of Neural Architectures in Remote Sensing Classification Problems}. In: 2021
  IEEE International Geoscience and Remote Sensing Symposium IGARSS. pp.
  1587--1590. IEEE (2021). \doi{10.1109/IGARSS47720.2021.9554309}

\bibitem{elsken2019neural}
Elsken, T., Metzen, J.H., Hutter, F.: {Neural Architecture Search: A Survey}.
  The Journal of Machine Learning Research  \textbf{20}(1),  1997--2017 (2019)

\bibitem{feurer2019hyperparameter}
Feurer, M., Hutter, F.: {Hyperparameter Optimization}. In: {Automated Machine
  Learning}, pp. 3--33. Springer, Cham (2019)

\bibitem{fortin2012deap}
Fortin, F.A., De~Rainville, F.M., Gardner, M.A.G., Parizeau, M., Gagn{\'e}, C.:
  {DEAP: Evolutionary Algorithms Made Easy}. The Journal of Machine Learning
  Research  \textbf{13}(1),  2171--2175 (2012)

\bibitem{funk2022prediction}
{Funk, Yannick and G{\"o}tz, Markus, and Anzt, Hartwig}: {Prediction of Optimal
  Solvers for Sparse Linear Systems Using Deep Learning}. In: Proceedings of
  the 2022 SIAM Conference on Parallel Processing for Scientific Computing. pp.
  14--24. Society for Industrial and Applied Mathematics (2022).
  \doi{10.1137/1.9781611977141.2}

\bibitem{george2020katib}
George, J., Gao, C., Liu, R., Liu, H.G., Tang, Y., Pydipaty, R., Saha, A.K.: {A
  Scalable and Cloud-Native Hyperparameter Tuning System} (2020).
  \doi{10.48550/arXiv.2006.02085}

\bibitem{golovin2017google}
Golovin, D., Solnik, B., Moitra, S., Kochanski, G., Karro, J., Sculley, D.:
  {Google Vizier: A Service for Black-Box Optimization}. In: Proceedings of the
  23rd ACM SIGKDD International Conference on Knowledge Discovery and Data
  Mining. pp. 1487--1495 (2017). \doi{10.1145/3097983.3098043}

\bibitem{he2016deep}
He, K., Zhang, X., Ren, S., Sun, J.: {Deep Residual Learning for Image
  Recognition}. In: Proceedings of the IEEE conference on computer vision and
  pattern recognition. pp. 770--778 (2016)

\bibitem{he2016identity}
He, K., Zhang, X., Ren, S., Sun, J.: {Identity Mappings in Deep Residual
  Networks}. In: European Conference on Computer Vision. pp. 630--645. Springer
  (2016). \doi{10.1007/978-3-319-46493-0_38}

\bibitem{hertel2018sherpa}
Hertel, L., Collado, J., Sadowski, P., Baldi, P.: {Sherpa: Hyperparameter
  Optimization for Machine Learning Models}. 32nd Conference on Neural
  Information Processing Systems (NIPS 2018)  (2018),
  \url{https://github.com/sherpa-ai/sherpa}

\bibitem{holland1992adaptation}
Holland, J.H.: {Adaptation in Natural and Artificial Systems: An Introductory
  Analysis with Applications to Biology, Control, and Artificial Intelligence}.
  MIT press (1992). \doi{10.7551/MITPRESS/1090.001.0001}

\bibitem{hutter2011sequential}
Hutter, F., Hoos, H.H., Leyton-Brown, K.: {Sequential Model-based Optimization
  for General Algorithm Configuration}. In: International Conference on
  Learning and Intelligent Optimization. pp. 507--523. Springer (2011).
  \doi{10.1007/978-3-642-25566-3_40}

\bibitem{koch2018autotune}
Koch, P., Golovidov, O., Gardner, S., Wujek, B., Griffin, J., Xu, Y.:
  {Autotune: A Derivative-free Optimization Framework for Hyperparameter
  Tuning}. In: Proceedings of the 24th ACM SIGKDD International Conference on
  Knowledge Discovery \& Data Mining. pp. 443--452 (2018).
  \doi{10.1145/3219819.3219837}

\bibitem{liaw2018tune}
Liaw, R., Liang, E., Nishihara, R., Moritz, P., Gonzalez, J.E., Stoica, I.:
  {Tune: A Research Platform for Distributed Model Selection and Training}.
  arXiv preprint arXiv:1807.05118  (2018). \doi{10.48550/arXiv.1807.05118}

\bibitem{lindauer2022smac3}
Lindauer, M., Eggensperger, K., Feurer, M., Biedenkapp, A., Deng, D.,
  Benjamins, C., Ruhkopf, T., Sass, R., Hutter, F.: Smac3: A versatile bayesian
  optimization package for hyperparameter optimization. J. Mach. Learn. Res.
  \textbf{23},  54--1 (2022)

\bibitem{lunacek2019impact}
Lunacek, M., Whitley, D., Sutton, A.: {The Impact of Global Structure on
  Search}. In: {Parallel Problem Solving from Nature -- PPSN X, LNCS 5199}. pp.
  498--507. Springer (2008). \doi{10.1007/978-3-540-87700-4_50}

\bibitem{luque2011parallel}
Luque, G., Alba, E.: {Parallel Genetic Algorithms: Theory and Real World
  Applications}, vol.~367. Springer (2011). \doi{10.1007/978-3-642-22084-5}

\bibitem{mitchell1998introduction}
Mitchell, M.: {An Introduction to Genetic Algorithms}. MIT Press (1998)

\bibitem{nevergrad}
Rapin, J., Teytaud, O.: {Nevergrad - A Gradient-free Optimization Platform}.
  \url{https://GitHub.com/FacebookResearch/Nevergrad} (2018)

\bibitem{snoek2012practical}
Snoek, J., Larochelle, H., Adams, R.P.: {Practical Bayesian Optimization of
  Machine Learning Algorithms}. In: Pereira, F., Burges, C., Bottou, L.,
  Weinberger, K. (eds.) Advances in Neural Information Processing Systems.
  vol.~25. Curran Associates, Inc. (2012),
  \url{https://proceedings.neurips.cc/paper/2012/file/05311655a15b75fab86956663e1819cd-Paper.pdf}

\bibitem{song2022open}
Song, X., Perel, S., Lee, C., Kochanski, G., Golovin, D.: {Open Source Vizier:
  Distributed Infrastructure and API for Reliable and Flexible Blackbox
  Optimization}. In: Automated Machine Learning Conference, Systems Track
  (AutoML-Conf Systems) (2022), \url{https://github.com/google/vizier}

\bibitem{sudholt2015pea}
Sudholt, D.: {Parallel Evolutionary Algorithms -- Chapter in the Handbook of
  Computational Intelligence}. Springer (2015).
  \doi{10.1007/978-3-662-43505-2_46}

\bibitem{sumbul2020bigearthnet}
Sumbul, G., Kang, J., Kreuziger, T., Marcelino, F., Costa, H., Benevides, P.,
  Caetano, M., Demir, B.: {BigEarthNet Dataset with a New Class-Nomenclature
  for Remote Sensing Image Understanding}. arXiv preprint arXiv:2001.06372
  (2020). \doi{10.48550/arXiv.2001.06372}

\bibitem{EvoTorch}
Toklu, N.E., Atkinson, T., Micka, V., Srivastava, R.K.: {EvoTorch: Advanced
  evolutionary computation library built directly on top of PyTorch, created at
  NNAISENSE.} https://github.com/nnaisense/evotorch (2022)

\bibitem{tomassini2006spatially}
Tomassini, M.: {Spatially Structured Evolutionary Algorithms: Artificial
  Evolution in Space and Time}. Springer (2006). \doi{10.1007/3-540-29938-6}

\bibitem{wang2020parallel}
Wang, J., Clark, S.C., Liu, E., Frazier, P.I.: {Parallel Bayesian Global
  Optimization of Expensive Functions}. Operations Research  \textbf{68}(6),
  1850--1865 (2020). \doi{10.1287/opre.2019.1966}

\bibitem{weiel2021dynamic}
Weiel, M., G{\"o}tz, M., Klein, A., Coquelin, D., Floca, R., Schug, A.: Dynamic
  particle swarm optimization of biomolecular simulation parameters with
  flexible ojective functions. Nature Machine Intelligence  \textbf{3}(8),
  727--734 (2021). \doi{10.1038/s42256-021-00366-3}

\bibitem{young2015optimizing}
Young, S.R., Rose, D.C., Karnowski, T.P., Lim, S.H., Patton, R.M.: {Optimizing
  Deep Learning Hyper-parameters through an Evolutionary Algorithm}. In:
  Proceedings of the Workshop on Machine Learning in High-performance Computing
  Environments. pp.~1--5 (2015). \doi{10.1145/2834892.2834896}

\end{thebibliography}
\end{document}